\documentclass[sigconf,nonacm]{acmart}

\usepackage{booktabs}
\usepackage{makecell}
\usepackage{multirow}
\usepackage[most]{tcolorbox}
\usepackage{diagbox}
\usepackage{ifthen}
\usepackage{soul}

\newboolean{highlight-changes}
\setboolean{highlight-changes}{false} 

\newcommand{\hladd}[1]{\ifthenelse{\boolean{highlight-changes}}{\textcolor[HTML]{19BC09}{#1}}{#1}}
\newcommand{\hlremove}[1]{\ifthenelse{\boolean{highlight-changes}}{\textcolor[HTML]{FF4545}{#1}}{}}
\newcommand{\hlreplace}[2]{\hlremove{#1}\hladd{#2}}

\definecolor{nicegreen}{rgb}{0.099,0.737,0.035}
\definecolor{nicered}{rgb}{1.000,0.271,0.271}
\definecolor{niceblue}{rgb}{0.271,0.510,1.000}
\definecolor{lightergreen}{rgb}{0.567,0.973,0.516}
\definecolor{niceorange}{rgb}{1.000, 0.678, 0.086}

\newtcbox{\todobadge}[1][niceorange]{
  on line,
  arc=1pt,
  colback=#1!90!black,
  colframe=#1!90!black,
  fontupper=\tiny\strut\color{white},
  boxrule=1pt, 
  boxsep=0pt,
  left=1pt,
  right=1pt,
  top=0.1pt,
  bottom=0.1pt
}

\newcommand{\ds}[1]{\textsc{\footnotesize#1}}

\newcommand{\textBF}[1]{%
    \pdfliteral direct {2 Tr 0.3 w} 
     #1%
    \pdfliteral direct {0 Tr 0 w}%
}

\newcommand{\tb}[1]{\textBF{#1}}

\begin{document}

\title{A Missing Data Imputation GAN for Character Sprite Generation}

\author{Fl\'{a}vio Coutinho}
\orcid{0000-0001-8014-3906}
\additionalaffiliation{%
    \institution{Centro Federal de Educação Tecnológica de Minas Gerais}
    \department{DECOM}
    \city{Belo Horizonte}
    \state{MG}
    \country{Brazil}
}
\affiliation{%
    \institution{Universidade Federal de Minas Gerais}
    \department{Departamento de Ciência da Computação}
    \city{Belo Horizonte}
    \state{MG}
    \country{Brazil}
}
\email{fegemo@cefetmg.br}

\author{Luiz Chaimowicz}
\orcid{0000-0001-8156-9941}
\affiliation{%
  \institution{Universidade Federal de Minas Gerais}
  \department{Departamento de Ciência da Computação}
  \city{Belo Horizonte}
  \state{MG}
  \country{Brazil}
}
\email{chaimo@dcc.ufmg.br}



\begin{abstract}
Creating and updating pixel art character sprites with many frames spanning different animations and poses takes time and can quickly become repetitive. However, that can be partially automated to allow artists to focus on more creative tasks. In this work, we concentrate on creating pixel art character sprites in a target pose from images of them facing other three directions. We present a novel approach to character generation by framing the problem as a missing data imputation task. Our proposed generative adversarial networks model receives the images of a character in all available domains and produces the image of the missing pose. We evaluated our approach in the scenarios with one, two, and three missing images, achieving similar or better results to the state-of-the-art when more images are available. We also evaluate the impact of the proposed changes to the base architecture.
\end{abstract}

\ccsdesc[300]{Computing methodologies~Computer vision}
\ccsdesc[500]{Computing methodologies~Computer graphics}
\ccsdesc[500]{Computing methodologies~Neural networks}

\keywords{Generative Adversarial Networks, Procedural Content Generation, Image-to-Image Translation, Missing Data Imputation, Character Sprites}

\begin{teaserfigure}
  \includegraphics[width=\textwidth]{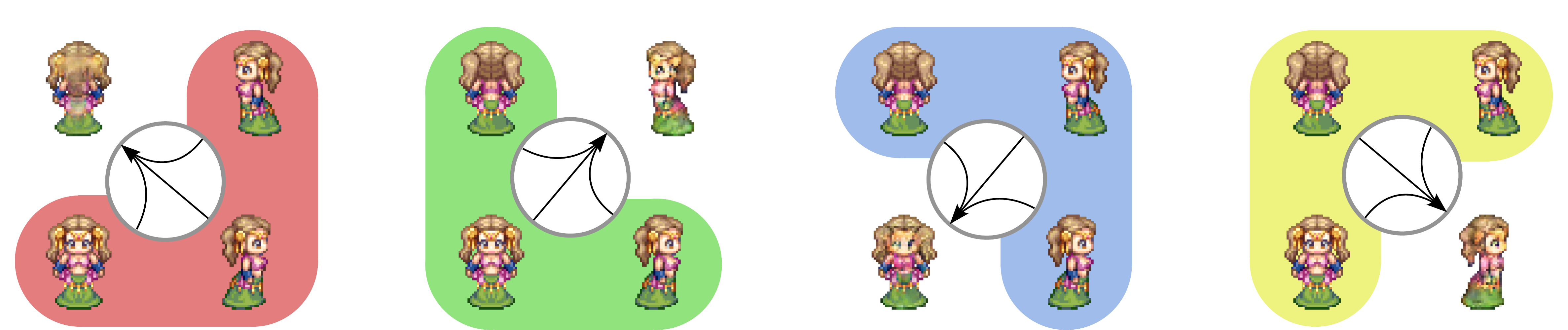}
  \caption{Our model imputes a character's missing pose collaboratively using all the available information from other domains.}
  \Description[Sample character with a missing pose generated using our proposed model]{There are four examples, each displaying three poses of a character (e.g., facing right, front, left) being used to generate a missing sprite (e.g., facing back). Each example uses the other three directions to produce a different missing pose.}
  \label{fig:teaser}
\end{teaserfigure}


\maketitle

\section{Introduction}
Asset creation is a vital part of the game development process, and it usually takes up a large portion of the project schedule. In particular, the task of character design is seldom executed in a forward-only way, typically involving a lot of going back and forth~\citep{Schreier2017BloodMade}. In pixel art games, in which the color of each pixel is thoughtfully picked, even small changes to a character might require updating many sprites, especially if characters can face multiple directions and contain different animation sequences spanning many frames~\citep{Silber2015PixelDevelopers}.

Despite the character creation process requiring high creativity and being an established and well-suited responsibility for artists, some involved tasks can become repetitive. For instance, creating normal maps~\citep{Moreira2022AnalysisArt} from colored sprites, designing every animation frame~\citep{Coutinho2022GeneratingGANs}, or propagating changes to the many sprites of a character. In that context, recent Procedural Content Generation techniques can help streamline the pipeline, particularly those involving Machine Learning (PCGML). Different works approached character generation through PCGML techniques using Variational Autoencoders (VAEs)~\citep{Loftsdottir2022SketchBetween:Sketches,Saravanan2022PixelRepresentation}, Generative Adversarial Networks (GANS)~\citep{Hong2019GameGAN,Coutinho2022GeneratingGANs,Coutinho2024PixelGANs,Serpa2019TowardsSheets,Choi2022ImageDrawing}, and Convolutional Neural Networks (CNNs)~\citep{Serpa2022HumanLearning}, and all of them posed their problems as an image to image translation task that generates an image given another (e.g., a normal map from a shaded character). However, if more information is available to the model, it can be leveraged to potentially generate better images.

In this work, we tackle the problem of generating a character sprite in a target pose as a missing data imputation task, using all the images of the character available in other poses. In particular, we propose a model that uses images of pixel art characters in source poses (e.g., facing left, right, back) to impute a missing target direction (e.g., facing front). Figure~\ref{fig:teaser} illustrates our approach.

We propose a generative adversarial network model based on the CollaGAN~\citep{Lee2019CollaGAN:Imputation} architecture, with changes to the generator topology and the training procedure. Compared to the baselines using the metrics Frechét Inception Distance (FID)~\citep{Heusel2017GANsEquilibrium} and $L_1$ distance, the images produced by our model are similar or better than the state-of-the-art. When fewer images are available, the model still produces feasible images, but with less quality. In an ablation study, we show how each of the proposed changes to the original CollaGAN influenced the improved results we achieved.

Thus, our main contributions in this work are:
\begin{itemize}
    \item a GAN with a single generator/discriminator that can target multiple character poses;
    \item empirical demonstration that using more of the available information improves the produced sprites; and
    \item changes to the CollaGAN architecture that enhance the quality of the generated images.
\end{itemize}

\section{Background}
In this section we describe some concepts related to generative adversarial networks and then present how such models can approach the image-to-image translation problem. 

\subsection{Generative Adversarial Networks}
Goodfellow et al. \citep{Goodfellow2014GenerativeNets} introduced the concept of generative adversarial networks (GANs) as a framework for generating content through an adversarial training process. It consists of two models playing different roles in a minimax game: a generator $G$ that evolves to create new content similar to the examples seen during training and a discriminator $D$ that learns to distinguish between real and generated (fake) examples. If an optimal state is reached, $G$ captures the distribution of the training data and can produce new samples that are indistinguishable from the real ones. At the same time, $D$ cannot tell whether an observation is real or fake.

The training algorithm traverses the set of examples for a number of epochs. At each step, $G$ receives a noise prior $z$ and produces new examples, while $D$ is called once to discriminate a minibatch of generated samples and a second time with real ones. The discriminator's loss function $\mathcal{L}_D$ is the mean value (of all examples) of (the log of) the probability of incorrectly labeling an input example $x$ as fake and (the log of) the probability of incorrectly labeling a fake example $G(z)$ as real. $D$ updates its weights \textit{ascending} the stochastic gradient of the following value function, while $G$ updates \textit{descending} it:
\begin{equation*}\label{equ:original-gan-value-function}
\min_G\max_D\mathcal{L}(G,D)=\mathbb{E}_{x}\lbrack\log{D(x)}\rbrack+\mathbb{E}_{z}\lbrack\log{(1-D(G(z)))}\rbrack
\end{equation*}

In this work, we propose a GAN model to generate pixel art character sprites. In particular, it generates a character in a target pose given input images of it facing other source directions. Because it uses multiple images as input to generate a missing one, we approach the problem as a missing data imputation task. However, it can also be regarded as an image-to-image translation problem with multiple images as input and the target as the missing domain. Next, we define the image-to-image translation task and present some of the proposed deep generative architectures using GANs.

\subsection{Image-to-Image Translation}
Pang et al. \cite{Pang2022Image-to-ImageApplications} define image-to-image translation as the process of converting an input image $x_a$ in a source domain $a$ to a target $b$  while keeping some intrinsic content from $a$ and transferring it to the extrinsic style of $b$. The meaning of a domain, style and content differ according to the task. To illustrate, if we want to create a cartoon version (domain $b$) from pictures of faces (domain $a$), we are translating faces $x_a$ to $b$, keeping the person's identity (intrinsic content) but using cartoonish techniques (extrinsic style). Different problems have been approached as image-to-image translation using deep generative models (e.g., GANs, VAEs), such as image colorization \citep{Jiang2021GAN-AssistedGeneration, Gonzalez2020GeneratingLearning}, semantic image synthesis \citep{Serpa2019TowardsSheets,Isola2017Image-to-ImageNetworks}, style transfer \citep{Zhu2017UnpairedNetworks}, attribute manipulation \citep{Choi2018StarGAN:Translation,Choi2020StarGANDomains,Lee2019CollaGAN:Imputation}, and pose transfer \citep{Hong2019GameGAN, Coutinho2022GeneratingGANs,Coutinho2024PixelGANs}. 

The diversity of the presented problems involves different characteristics of the task and the proposed solution. A first important property is the use of \textbf{supervision} (label/annotated examples) for training, which largely depends on the availability of such data. For instance, in a translation from grayscale to colored pictures, it is easy to have pixel-wise aligned examples, but that is not the case if we want to transform horses into zebras, as the cost of acquiring completely registered pairs of photos of horses and zebras in the same position in the same environment is impractical. Hence, when paired data is available for some task, we can use \textbf{supervised} training~\citep{Isola2017Image-to-ImageNetworks,Lee2019CollaGAN:Imputation}, whereas when it is not, the algorithm needs to train in an \textbf{unsupervised} fashion~\citep{Zhu2017UnpairedNetworks,Choi2018StarGAN:Translation,Afifi2021HistoGAN:Histograms}.

A second characteristic of the tasks is the \textbf{number of domains} involved in the translation and how the proposed architecture can deal with them. For instance, many problems consist of only two domains (e.g., grayscale to color, photo to painting, semantic labels to photographs). In contrast, others involve multiple (e.g., translating a neutral face to one smiling, angry, or crying). Hence, the proposed architectures can be \textbf{two-domain} \citep{Isola2017Image-to-ImageNetworks,Zhu2017UnpairedNetworks} or \textbf{multi-domain}, supporting the translation among all directions \citep{Choi2018StarGAN:Translation,Choi2020StarGANDomains,Lee2019CollaGAN:Imputation}. Additionally, the architectures for two-domain translation can generate images in a single direction \citep{Isola2017Image-to-ImageNetworks} or in both \citep{Zhu2017UnpairedNetworks,Zhu2017TowardTranslation}.

Authors have proposed architectures for tasks with different sets of characteristics. Here, we propose a model based on the Collaborative GAN (CollaGAN) to generate missing poses of pixel art characters. In our experiments, we compare the proposed architecture to baselines consisting of models based on Pix2Pix~\citep{Coutinho2024PixelGANs} and StarGAN~\citep{Choi2018StarGAN:Translation}.

Pix2Pix trains with supervision (paired images). It can translate images from one domain into another in a single direction. In contrast, StarGAN trains unsupervisedly but supports multiple domains with a single generator and discriminator pair. CollaGAN, in turn, requires supervision and is multi-domain, with the additional difference that it uses images from multiple domains as input.

\section{Related Work}
As we investigate the generation of character sprites, we first describe some recent works that tackle the automatic creation of characters. Most also deal with pixel art imagery and use deep generative models. In sequence, we present some works related to the missing data imputation problem, which is how we frame the generation of missing character poses.

\subsection{Sprite Generation}

Some works propose generating characters in a target pose using a bone graph to indicate the desired positions of each body part. Hong et al.~\citep{Hong2019GameGAN} approached that task with a multiple discriminator GAN~(MDGAN). It translates images of a character (representing its shape and color) and a target bone-graph sprite into a target of that same character in the new pose. The model consists of a generator and two discriminators, one to determine if two images share the same color and shape, while the other tells whether a character's pose is correct according to some bone-graph sprite.

Similarly, Choi et al.~\citep{Choi2022ImageDrawing} created a database of character sprites in walking and running animations by feeding video frames of real people into body segmentation networks. Then, they trained a model to generate characters from the body-segmented sprites in arbitrary poses created by users. Albeit successful in their proposed experiments, both systems require tailored datasets that match the positions of characters, making it challenging for the models to generalize, especially for games with different character shapes and movements. In addition, both works use real images in their training sets, which do not conform to characters in typical 2D games, especially those in the pixel art style.

Targeting the generation of in-between frames of animated sketches, Loftsdóttir and Guzdial~\citep{Loftsdottir2022SketchBetween:Sketches} propose SketchBetween: a model that takes the initial and final frames of a character animation and sketches of the internal frames, and generates colored versions of the frames in the middle. It takes five images of an incomplete animation as input and provides five images with the rendered sprite animation. Trained on a dataset of cartoon animal animations, it had promising results on shapes and poses similar to the ones from the training set. However, even the higher-quality examples presented the blurriness typical of how VAEs optimize to reduce the average reconstruction error.

Regarding the generation of pixel art characters, researchers approached differently: adding specific layers to the generator~\citep{Saravanan2022PixelRepresentation}, framing the problem as a semantic segmentation task~\citep{Serpa2022HumanLearning,Coutinho2022OnGANs}, doing post processing steps~\citep{Coutinho2024PixelGANs}, or adding a histogram loss~\citep{Coutinho2022OnGANs}.

Serpa and Rodrigues~\citep{Serpa2019TowardsSheets} proposed a model based on Pix2pix~\citep{Isola2017Image-to-ImageNetworks} to generate a grayscale-shaded sprite and another one that segments characters' body parts from rough line art sketches of animation frames from a fighting game. The generated grayscale sprites were close to the ground truth, but the colored ones diverged, especially for characters in less common poses. In a later iteration of the work~\cite{Serpa2022HumanLearning}, the authors got improved results by framing the problem as a semantic segmentation task and changing the architecture accordingly. The proposed model dropped the adversarial training and employed dense connections to increase the network's depth, deep supervision to provide gradients to every step, and a class-weighted focal loss to overcome the class imbalance in the training data.

Saravanan and Guzdial~\citep{Saravanan2022PixelRepresentation} adapted the VQ-VAE~\citep{vandenOord2017NeuralLearning} to improve the quality of the generated pixel art characters by adding a $1\times1$ convolution layer pair at the beginning and end of the encoder and decoder networks. Trained with Pokémon sprites, the model generated embeddings that allowed a PixelCNN~\citep{vandenOord2016PixelKavukcuoglu} technique to create new images of static characters that tried to follow the training distribution. Using the additional layers helped reduce the blurriness of the generated images.

Investigating the challenges involved in generating pixel art specifically, Coutinho and Chaimowicz~\citep{Coutinho2022OnGANs} evaluated two hypotheses: representing images as indices in a color palette and adding a histogram loss term when training the generator. While the palette representation led to much worse results due to overfitting, penalizing the generator for using colors with a different histogram than the one from the input image yielded slightly improved images.

In \citep{Coutinho2022GeneratingGANs}, the same authors propose an architecture based on Pix2Pix to translate pixel art characters in a source pose (e.g., looking front) into a target one (e.g., facing right). They trained models in different datasets with under 1k examples. The generated images had varying degrees of quality, with good results for characters more similar to the ones seen during training (e.g., similar shapes or color variations) but bad results for more unique characters. In a later iteration of the work, Coutinho and Chaimowicz~\citep{Coutinho2024PixelGANs} investigated different data augmentation techniques. They proposed a post-processing step to quantize the images to the color palette of the input image. They also assembled a diverse dataset with 14k paired images of characters in four directions and observed that training with much more data yielded better results when validating with the more artistically cohesive individual datasets.

In this paper, we also tackle the generation of pixel art characters (like \citep{Serpa2019TowardsSheets,Serpa2022HumanLearning,Saravanan2022PixelRepresentation}) by translating among different poses (like \citep{Coutinho2022OnGANs,Coutinho2022GeneratingGANs,Coutinho2024PixelGANs}). However, unlike the other works, we frame the problem as a missing image data imputation task. Hence, instead of a trained model being capable of handling between only two poses (two-domain) and a single direction, our generator receives the images of a character in every available pose and generates an image of it in the one that is missing (multi-domain\hladd{ with multiple inputs}).

\subsection{Missing Data Imputation}

Data analysis can be drastically hindered when relevant parts of information are missing. That can happen for various reasons: data can be \hlreplace{missing}{absent} because it was never collected or produced, it might have been lost, or it might contain errors~\cite {Yoon2018GAIN:Nets}. Researchers have proposed different missing data imputation techniques to replace absent data with plausible \hlreplace{replacements}{substitutions}. The choice of such techniques depends on the data type, among other characteristics. \hlreplace{The data type}{It} can be one or a mix of categorical~\citep{Yoon2018GAIN:Nets,Shang2017VIGAN:Networks}, sequential~\citep{Liu2023OneImputation}, and image~\citep{Shang2017VIGAN:Networks,Lee2019CollaGAN:Imputation,Sharma2019MissingNetwork,Shen2021Multi-DomainData}.

Inaugurating the use of deep learning-based techniques for missing data imputation, Yoon et al.~\citep{Yoon2018GAIN:Nets} proposed a generalization of the original GAN to deal with imputing missing values, which they called Generative Adversarial Imputation Nets (GAIN). The generator receives three inputs: the sample with missing values, a mask indicating which values are present, and a random vector of the same dimension that introduces noise. As output, it produces a version of the sample with replaced values for those missing. The discriminator, in turn, tries to distinguish which of the categorical variables are imputed and which are from the original sample.

The imputation task becomes more challenging when the missing data are images due to the higher dimensionality. Some works approach the problem using GANs~\citep{Shang2017VIGAN:Networks,Lee2019CollaGAN:Imputation,Sharma2019MissingNetwork,Shen2021Multi-DomainData}. An example is the View Imputation GAN (VIGAN)~\citep{Shang2017VIGAN:Networks}, that can generate missing values in a target domain by combining a modified CycleGAN~\citep{Zhu2017UnpairedNetworks} with a Denoising Autoencoder in a three-step training process. A shortcoming of VIGAN is that it performs bi-directional imputation between only two domains. When the task involves more domains, other architectures are better suited. The Multi-Modal GAN (MM-GAN)~\citep{Sharma2019MissingNetwork}, CollaGAN~\citep{Lee2019CollaGAN:Imputation}, and ReMIC~\citep{Shen2021Multi-DomainData} can impute missing images among multiple domains and use the information of all available sources as input to the generator.

MM-GAN's generator~\citep{Sharma2019MissingNetwork} has an equal number of inputs and outputs, receives samples with images missing in random domains, and outputs imputed values. The discriminator distinguishes between real and imputed same-size patches of a full sample comprising all domains. CollaGAN~\citep{Lee2019CollaGAN:Imputation} works similarly and was proposed in the same year as MM-GAN. However, it produces an image of a single target domain. Its generator varies depending on the task, but it also receives the images in all available domains, concatenated with the index of the target domain spread spatially and through the channels dimension. Both architectures presented good results in their respective experiments. ReMIC~\citep{Shen2021Multi-DomainData} also takes the inputs from all available domains and generates the missing ones, like MM-GAN. However, unlike the other two, it disentangles the images and extracts a shared content encoding and a separate style encoding for each domain. 

All multi-domain architectures that deal with missing image data imputation~\citep{Sharma2019MissingNetwork,Lee2019CollaGAN:Imputation,Shen2021Multi-DomainData} were tested either with medical or natural images, but not with pixel art or other \hlreplace{forms of art}{styles}. In the next section, we present a modified architecture based on CollaGAN to generate missing pixel art characters.

\section{Architecture}

\begin{figure*}[t]
    \centering
    \includegraphics[width=1\linewidth]{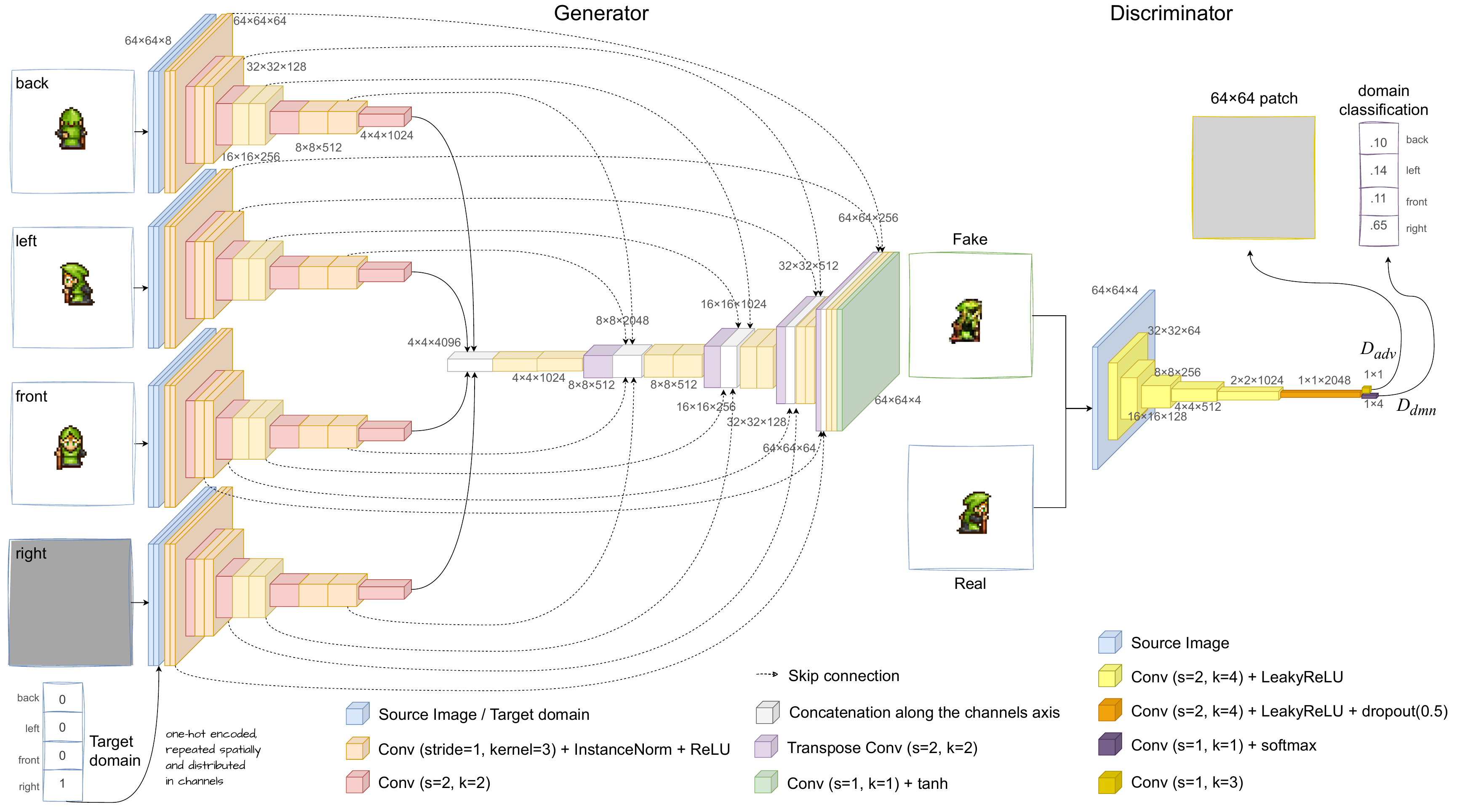}  
    \caption{Architecture of the proposed model. \textsc{Left}: The generator receives a character in the source domains and a label indicating the target, which is one-hot encoded, spatially spread, and concatenated with each input image. The inputs follow the encoder branches and are concatenated at the bottleneck layer, flowing into the unified decoder. Skip connections provide early outputs to the decoder. \textsc{Right}: The discriminator receives the image (real or fake) that must be distinguished and outputs $D_{adv}$ with the real/fake logit and $D_{dmn}$ with the probabilities of the image being part of each domain.}
    \label{fig:architecture-collagan}
    \Description[Architecture diagram of the proposed model]{Two neural networks, the generator on the left and the discriminator on the right, are depicted by the successive activation maps and skip connections. The generator has four inputs that go into a separate encoder branch that progressively reduces the image dimension while increasing the number of channels until a 4 by 4 by 2058 size. Then, a single decoder upscales the resolution while decreasing the depth until an output image is generated with 64 by 64 by 4 channels. The discriminator takes either a generated or a real image as input, reduces the dimensions while increasing the depth, and outputs both a value with the probability of the image being real and a classification of which domain the image comes from.}
\end{figure*}

We propose an architecture based on CollaGAN~\citep{Lee2019CollaGAN:Imputation} to impute images of pixel art characters in a missing pose (target domain). Considering that there are domains $N=\{a,b,c,d\}$, one representing each pose. The architecture consists of a single generator and discriminator pair that creates an image $\hat{x}_t$ of a character in the missing pose $t$ using the available images from all of the other source $S$ poses:
$$\hat{x}_t=G(x_S,t), \text{with } t\in N, S=N-\{t\}$$
Our generator has one encoder branch to process the input for each domain, a single decoder branch with concatenated skip connections, and outputs an image in the missing domain. The discriminator distinguishes images as real or fake, as well as determines their domain through an auxiliary classifier output. Figure~\ref{fig:architecture-collagan} shows the topology of both networks.

\subsection{Objective Function}
As usually done with GANs, we train both networks adversarially, but also with additional objectives. The generator's loss function has five terms: regressive, cycle consistency, structural similarity, adversarial, and domain classification. In turn, the discriminator trains with adversarial and domain classification objectives.

Training requires a forward and a backward pass. In the first step, a minibatch of paired images with a random missing domain $t$ is fed to the generator $G$, which synthesizes an image corresponding to the missing $t$ domain. For example, if $S=\{a,b,c\}$ and $t=d$, the images $x_a$, $x_b$, $x_c$ are available and we want the model to generate $\hat{x}_d$ as close as possible to the real $x_d$:
$$\hat{x}_d=G(\{x_a,x_b,x_c,x_{\text{zero}}\}, d),$$
in which $x_{\text{zero}}$ is a tensor filled with zeros.

Subsequently, to ensure cycle consistency, the backward step comprises synthesizing $|N|-1$ images with each domain in $S=\{a,b,c\}$ as a target, using the generated $\hat{x}_d$ instead of the real $x_d$. The outputs of this pass, in our example, would be:
\begin{align*}
    \tilde{x}_{a|d}=&G(\{x_{\text{zero}},x_b,x_c,\hat{x}_d\}, a)\\
    \tilde{x}_{b|d}=&G(\{x_a,x_{\text{zero}},x_c,\hat{x}_d\}, b)\\
    \tilde{x}_{c|d}=&G(\{x_a,x_b,x_{\text{zero}},\hat{x}_d\}, c),
\end{align*}
and should reconstruct the original images $x_a$, $x_b$, and $x_c$.

A regressive loss term $\mathcal{L}_{reg}$ steers the generator towards using the information from the source domains to translate an image to the target\hlremove{ domain}, whereas a multiple cycle consistency loss $\mathcal{L}_{mcyc}$ leads it into encoding in $\hat{x}_t$  enough information to allow cyclical reconstruction of the original inputs. Both losses are pixel-wise $L_1$ distances between the generated and the real images:
\begin{align*}
    \mathcal{L}_{reg}=\mathbb{E}_{x_t,x_S}&[\lVert x_t-\hat{x}_t\rVert_1]\\[0.8em]
    \mathcal{L}_{mcyc}=\mathbb{E}_{x_t,x_S}&[\sum_{s\in S}\lVert x_s-\tilde{x}_{s|t}\rVert_1] 
\end{align*}

Besides $\mathcal{L}_{mcyc}$, an additional objective $\mathcal{L}_{ssim}$ is used to improve the quality of the images generated in the backward pass. It uses the structural similarity index measure (SSIM)~\citep{Wang2004ImageSimilarity}  to compose a loss term between the cyclically generated $\tilde{x}_S$ and the real source images $x_S$. Its formulation is the same as in the CollaGAN paper and is omitted here for brevity.

The discriminator also uses the other two objectives for the generator: adversarial and domain classification. The adversarial loss uses the one from Least Squares GAN~\citep{Mao2017LeastNetworks}, which optimizes the square of the errors of the discriminator classification of real and fake images. The discriminator  $\mathcal{L}_{adv}^D$ and generator  $\mathcal{L}_{adv}^G$ adversarial losses are: 
\begin{align*}
    \mathcal{L}_{adv}^D=&\mathbb{E}_{x_t}[(D_{adv}(x_t)-1)^2]+\mathbb{E}_{\tilde{x}_{s|t}}[(D_{adv}(\tilde{x}_{s|t}))^2]\\[.5em]
    \mathcal{L}_{adv}^G=&\mathbb{E}_{\tilde{x}_{s|t}}[(D_{adv}(\tilde{x}_{s|t})-1)^2]
\end{align*}
The domain classification objective leads the generator to synthesize images classified as having the intended target domain. For the generator, $\mathcal{L}_{dmn}^{fake}$ considers only generated images, whereas for the discriminator, $\mathcal{L}_{dmn}^{real}$ uses only real images. As a classification, they are calculated using cross entropy, given as:
\begin{align*}
    \mathcal{L}_{dmn}^{real}=&\mathbb{E}_{x_t}[-\log(D_{dmn}(x_t))]\\[0.5em]
    \mathcal{L}_{dmn}^{fake}=&\mathbb{E}_{\hat{x}_{t}}[-\log(D_{dmn}(\hat{x}_t))]
\end{align*}
To summarize, the full objectives of the generator $\mathcal{L}_G$ and the discriminator $\mathcal{L}_D$ are sums weighted by $\lambda$ scalars given as:
\begin{align*}
    \mathcal{L}_G=&\mathcal{L}_{adv}^G+\lambda_{reg}\mathcal{L}_{reg}+\lambda_{mcyc}\mathcal{L}_{mcyc}+\lambda_{ssim}\mathcal{L}_{ssim}+\lambda_{dmn}\mathcal{L}_{dmn}^{fake}\\[0.5em]
    \mathcal{L}_D=&\mathcal{L}_{adv}^D+\lambda_{dmn}\mathcal{L}_{dmn}^{real}
\end{align*}

\subsection{Generator}
The generator has four encoder branches, each receiving a source image from a particular domain and a channelized and spatially spread one-hot encoded label of the target domain. There are four downsampling blocks for each branch and a bottleneck layer that concatenates the activation maps from all encoder branches and further processes it. The data is then passed onto a single decoder composed of four upsampling blocks. There are skip connections from the concatenated activation maps (across the encoder branches) from downsampling to the respective upsampling blocks. Compared to the original architecture, our generator contains four encoder branches, while theirs has eight.

\begin{figure*}[t]
    \centering
    \includegraphics[width=1.0\linewidth,interpolate=false]{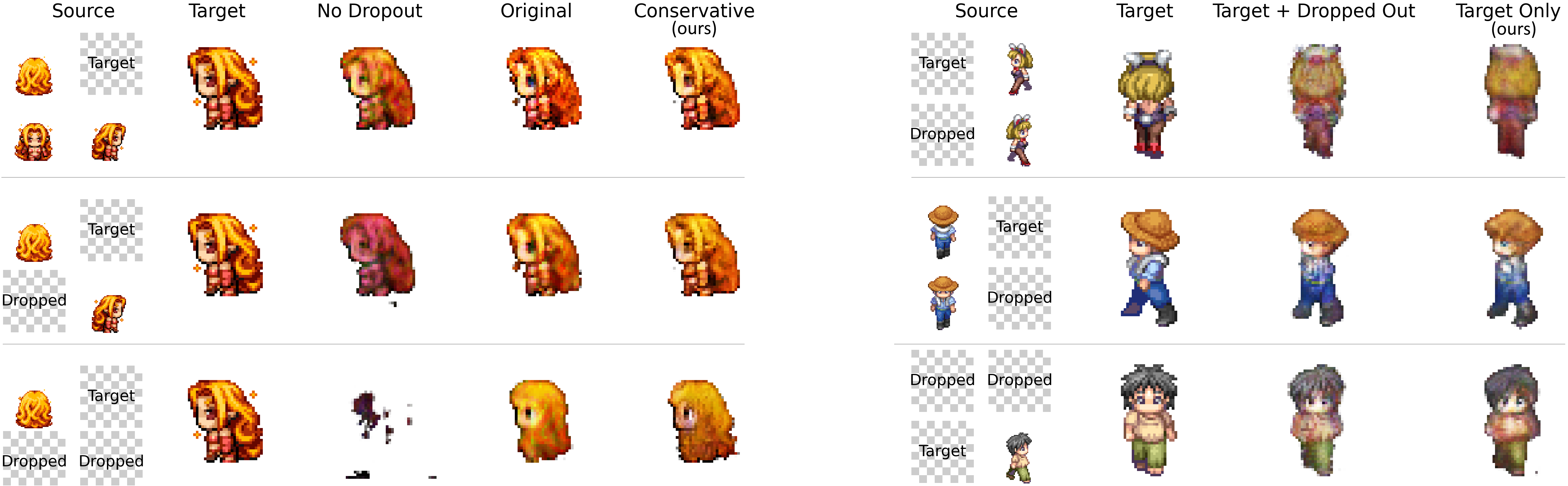}
    \caption{Comparison of input dropout (left) and replacement procedures (right) during training in the proposed model.}
    \Description[Examples of a character generated with different input dropout training strategies in scenarios of three input images, two, and one, and characters generated with the replacement procedures of the target + dropped out images and only the target]{On the left is a grid of three rows and five columns. The first row shows a character generated using three input images, the second with two, and the third with a single image. In the columns, the first shows the input images, the second is the target, the third to the fifth show the generated image of the character using no input dropout, the original dropout strategy from CollaGAN, and our proposed conservative input dropout tactic. The quality of the generated images degrades with fewer images as input, but it is much worse when no dropout is used. On the right side, three rows, each containing an example character generated with less than three images as input, using models trained with the replacement strategy of substituting both the target image and the dropped out and replacing only the target image. The images generated by the former model contain artifacts from domains other than the target (e.g., we can see two eyes in a target facing left, which is wrong).}
    \label{fig:collagan-dropout-and-replacer}
\end{figure*}

The image size and number of channels we use is $64\times64\times4$\hlremove{ like in the other proposed models}, contrasted to the $128\times128\times3$ configuration of the original architecture. We \hlreplace{used a higher}{increased the} number of channels for each layer to \hlreplace{increase}{improve} the network capacity: they are four times the original, becoming 64, 128, 256, 512, and 1024 for the blocks in each encoder branch, and 1024, 512, 256, 128, and 64 for the decoder blocks.

\subsection{Discriminator}
The discriminator receives a batch of images and outputs values $D_{adv}$ that should be one for real images and zero for the generated ones. In addition, it classifies the domain of the image, yielding probabilities $D_{dmn}$ of images having each domain.

The network topology is the same as in the original, with 6 downsampling blocks, each consisting of a convolution that halves the resolution while increasing the number of channels, with a leaky ReLU activation. The last block also contains a dropout layer. Following it, two parallel convolutions represent the $D_{adv}$ and the $D_{dmn}$ outputs, with linear and softmax activations, respectively.

\subsection{Training Procedure}

At each training step, we select a batch of paired images $x_S$ with random target domains $t$. The generator receives the batch of $\langle x_S,t\rangle$ and creates the missing $\hat{x}_t$, in the forward pass. Next, $\hat{x}_t$ is used in place of $x_t$ to create a number of new batches equal to $|N|-1$, in which each domain in $S$ becomes the target, in the backward (or cyclical) pass. The generator then creates $\tilde{x}_{s|t}, \forall s \in S$ that must be as close as possible to the original $x_s, \forall s \in S$.


The CollaGAN architecture authors observed that images are much worse as the number of available sources decreases. However, it is common to have use cases in which more than one domain is missing. Hence, they proposed a batch selection strategy called input dropout, in which the model trains with one or more missing domains. For instance, for $|N|=4$ and $t=d$, when a batch $\langle x_S,t\rangle$ is selected using the input dropout strategy, $x_S$ can have zero, one or two withdrawn images and be one of the following:
\begin{alignat*}{2}
    x_S=\{ x_a, x_b, x_c\}\quad&x_S=\{ x_{zero}, x_b, x_c\}\quad&&x_S=\{ x_{zero}, x_{zero}, x_c\}\\
    &x_S=\{ x_a, x_{zero}, x_c\}&&x_S=\{ x_{zero}, x_b, x_{zero}\}\\
    &x_S=\{ x_a, x_b, x_{zero}\}&&x_S=\{ x_a, x_{zero}, x_{zero}\}
\end{alignat*}

\hlreplace{The}{In the original CollaGAN, the} number of images to be dropped out is chosen uniformly, leaving a $33\%$ chance of having the full source domain set. That strategy improved the results in our task too. However, we observed that a more conservative approach in which the model trains more frequently dropping out few images yields even better results in the scenario of having fewer available images. We adopted chances of $10\%$, $30\%$, and $60\%$ to have two, one, and zero images dropped out. Figure~\ref{fig:collagan-dropout-and-replacer} (left) compares the three strategies (no dropout, original dropout, conservative dropout) with different numbers of missing images.

Another change we made to the training procedure relates to the backward generation pass. When the cycled images $\tilde{x}_{s|t}, \forall s \in S$  (e.g., $\tilde{x}_{a|d}$, $\tilde{x}_{b|d}$, and $\tilde{x}_{c|d}$) are generated in the original implementation, the image $\hat{x}_t$ generated in the forward pass replaces not only the original target image $x_t$, but also all images that have been dropped out due to the batch selection strategy. We experimented with having $\hat{x}_t$ replace only $x_t$ and observed better results.

To illustrate the difference, considering a batch with $t=d$ and the domain $c$ dropped out, the backward generated images $\tilde{x}_{s|t}, \forall s\in S$ for the original (left) and our implementation (right) would be as shown next. We highlighted the differences in color:
\begin{alignat*}{2}
    \tilde{x}_{a|d}=&G(\{x_{\text{zero}},x_b,\mathcolor{niceblue}{\hat{x}_d},\hat{x}_d\}, a)\qquad
    \tilde{x}_{a|d}=&&G(\{x_{\text{zero}},x_b,\mathcolor{niceblue}{x_{\text{zero}}},\hat{x}_d\}, a)\\
    \tilde{x}_{b|d}=&G(\{x_a,x_{\text{zero}},\mathcolor{niceblue}{\hat{x}_d},\hat{x}_d\}, b)\qquad
    \tilde{x}_{b|d}=&&G(\{x_a,x_{\text{zero}},\mathcolor{niceblue}{x_{\text{zero}}},\hat{x}_d\}, b)\\
    \tilde{x}_{c|d}=&G(\{x_a,x_b,x_{\text{zero}},\hat{x}_d\}, c)\qquad
    \tilde{x}_{c|d}=&&G(\{x_a,x_b,x_{\text{zero}},\hat{x}_d\}, c)
\end{alignat*}
Figure~\ref{fig:collagan-dropout-and-replacer} (right) compares the generated images when the model trains using the original replacement procedure for the dropped-out images versus our version where only the forward target image is replaced by the one generated in the forward pass. We can note that with the original procedure, the generator produces images with artifacts from domains other than the target one, mostly noticeable through the wrong number of eyes in the examples.


Regarding the number of trainable parameters, the generator contains 104,887,616 values, and the discriminator has 44,726,272. After training, the generator takes \textasciitilde116 ms to produce an image using a GeForce GTX 1050 GPU.

\section{Methodology}
We start by presenting the datasets used in the experiments to propose and evaluate models for translating pixel art characters in different poses. Next, we describe the metrics $L_1$ and FID used to analyze the quality of the generated images using each model. Finally, we conclude the section by presenting the baseline models used in the experiments.

\subsection{Dataset}
\begin{figure}[t]
    \centering
    \includegraphics[width=1\linewidth,interpolate=false]{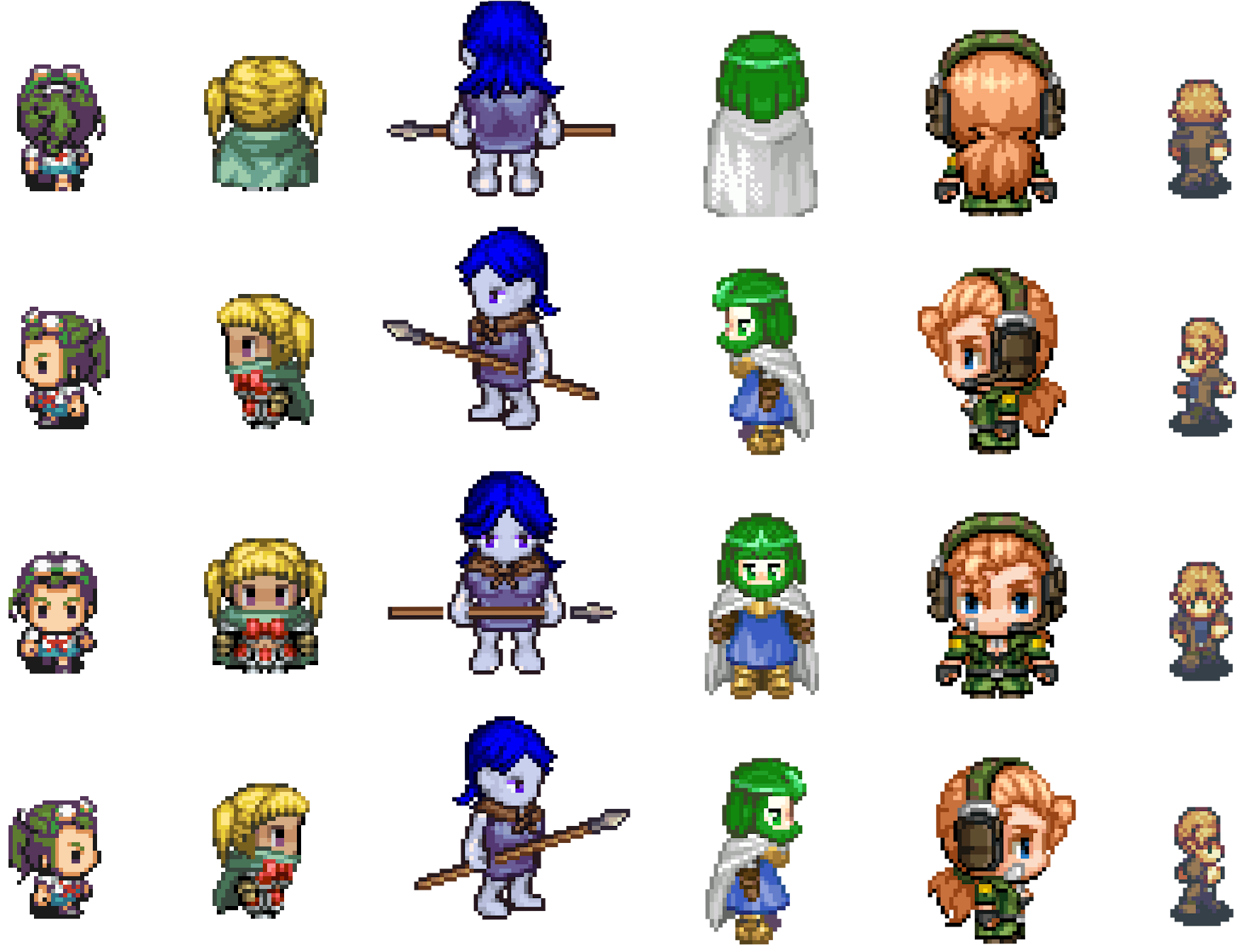}
    \caption{Sample images from the dataset showing different sizes/art styles (\textbf{columns}) facing four directions (\textbf{rows}).}
    \label{fig:datasets-samples}
    \Description[Example characters from the dataset]{There are six example characters (columns) facing back, left, front, and right (rows). Each character has a different size and art style, all in pixel art.}
\end{figure}

Unlike tasks that are more commonly tackled in Computer Vision research, we found only one character sprite dataset readily available: \ds{Tiny Hero}\footnote{Dual license of GNU GPL 3.0 and CC-BY-SA 3.0. Source: https://lpc.opengameart.org/}, which contains 912 paired images of characters facing the back, left, front, and right directions. To increase the number of training examples, we scraped character sprite sheets from different sources from the web, splitting them into individual character sprites, and generated characters modularly by assembling various parts. The dataset contains 14,202 paired images of characters in four directions spanning different art styles. They primarily comprise  humanoid characters of different sizes and art styles, but also a few sprites of animals, vehicles, and monsters. Figure~\ref{fig:datasets-samples} shows examples depicting the high variability of the samples. 

Images from each source had different character sizes, so the smaller ones were transparency-padded to the largest size, 64$\times$64. We also created an alpha channel with the character shape for the images that lacked one. The training set contains 12,074 examples, and the test set contains 2,128 examples (85\% split). \hladd{During training, we applied hue rotation to each character as data augmentation.}

\subsection{Evaluation Metrics}
The evaluation of generative models is an active research problem with different metrics proposed over the recent years \citep{Buzuti2023FrechetNetworks}. We evaluate the quality of a model by how close the generated images are to their ground truth. However, a qualitative analysis is important as the metrics do not always converge.

Hence, we analyze the results qualitatively through visual inspection and quantitatively using the $L_1$ distance to the target images and the Fréchet Inception Distance (FID) \citep{Heusel2017GANsEquilibrium}. The $L_1$ distance measures the absolute difference between the colors of pixels of two image sets (the generated and target). In turn, FID uses the Inception v3 network (proposed for image classification) to get the distance between the feature vectors of the two image sets~\citep{Szegedy2015RethinkingVision}. As both metrics are distances, they are zero for identical generated and target images, so lower numbers are better.

\subsection{Baseline Models}
We compare our model with two other architectures proposed for the image-to-image translation task: Pix2Pix~\citep{Isola2017Image-to-ImageNetworks} and StarGAN~\citep{Choi2018StarGAN:Translation}.

\textbf{Pix2Pix}. We trained a modified version of the architecture proposed in~\citep{Coutinho2024PixelGANs} for generating pixel art characters in a target pose given an image of it in a source one. Differently from the referenced work, we use 12 such models to support translation from and to all four poses: back, left, right, and front, excluding models from and to the same direction. Each generator has 29,307,844 trainable variables, so the model collection contains 351,694,128 parameters.

\textbf{StarGAN}. We trained a StarGAN-based model to perform multi-domain translation using a single generator and discriminator pair. The generator typically receives the source image and a label indicating the target direction. Still, we found that providing a label of the \textit{source} domain increases the quality of the generated characters. In turn, the original critic receives only the image to be evaluated, but we got better results by sending the source image too (before translation), which makes it perform a \textit{conditional} discrimination. In that case, the network indicates whether the provided image is real/fake \textit{considering that} it is a translation of the source image. For a fair comparison, we train the model using supervision (the original trains without paired images). The generator contains \hlreplace{8,446,208}{134,448,128} parameters.

\section{Experiments}
The model trained with the pixel art characters dataset for 240,000 generator update steps in minibatches of 4 examples, which is equivalent to \textasciitilde80 epochs. \hladd{It took 01:20h to train using a GeForce GTX 1050 GPU.} We used early stopping to select the model that had the best metrics on its test set instead of getting the one in the end to prevent overfitting. At every 1,000 update steps, we evaluate the model and select the one with the lowest (best) $L_1$ value throughout the training procedure. \hladd{After training, it takes 110.03ms for the model to generate a batch of images.}

The generator and discriminator optimize their weights using Adam, with $\beta_1=0.5$ and $\beta_2=0.999$, and a learning rate that starts as $0.0001$ and linearly decays to zero during the second half of the training. The parameters of the objective function were $\lambda_{reg}=100$, $\lambda_{dmn}=10$, $\lambda_{ssim}=10$, and $\lambda_{mcyc}=10$.

In the following experiments, we start by evaluating the model's performance using three input images (dubbed CollaGAN-3) against the baselines. Next, we evaluate the same model (trained with three source domains) in the scenario of it receiving only two (CollaGAN-2) and one (CollaGAN-1) input images. We then follow \hladd{up} with an experiment to assess different input dropout strategies and an ablation study of the changes proposed atop the original CollaGAN.


\subsection{Missing Image Imputation}
We trained our proposed model using conservative input dropout and the forward-only replacer strategy.
Tables~\ref{tab:baseline-comparison-fid} and \ref{tab:baseline-comparison-l1} show
the values of FID and $L_1$ for our proposed model and the baselines, with the rows representing the target pose and the columns displaying the metrics for the baselines Pix2Pix and StarGAN, averaged considering the translation from the other source domains and CollaGAN with the three other domains as input.

\begin{table}[t]
\centering
\caption{FID of our CollaGAN-based model receiving three images and a single for Pix2Pix and StarGAN}
\label{tab:baseline-comparison-fid}
\begin{tabular}{@{}lrrr@{}}
\toprule
\multirow{2}{*}{Target} & \multicolumn{3}{c}{Average FID}           \\ \cmidrule(l){2-4}
                        & Pix2Pix   & StarGAN   & CollaGAN-3        \\ \midrule
Back                    & 5.788     & 3.378     & \tb{2.054}    \\
Left                    & 2.380     & 1.250     & \tb{1.037}    \\
Front                   & 5.392     & 3.156     & \tb{1.955}    \\
Right                   & 2.806     & 1.368     & \tb{0.987}    \\ \midrule
Average                 & 4.091     & 2.288     & \tb{1.508}    \\ \bottomrule
\end{tabular}
\end{table}

\begin{table}[t]
\centering
\caption{$L_1$ of our CollaGAN-based model receiving three images and a single for Pix2Pix and StarGAN}
\label{tab:baseline-comparison-l1}
\begin{tabular}{@{}lrrr@{}}
\toprule
\multirow{2}{*}{Target} & \multicolumn{3}{c}{Average $L_1$}     \\ \cmidrule(l){2-4}
                        & Pix2Pix   & StarGAN   & CollaGAN-3    \\ \midrule
Back                    & 0.05402   & 0.06429   & \tb{0.04530}  \\
Left                    & 0.04934   & 0.06344   & \tb{0.03439}  \\
Front                   & 0.05875   & 0.07263   & \tb{0.04985}  \\
Right                   & 0.04880   & 0.06273   & \tb{0.03360}  \\ \midrule
Average                 & 0.05273   & 0.06577   & \tb{0.04078}  \\ \bottomrule
\end{tabular}
\end{table}


\begin{figure*}[t]
    \centering
    \includegraphics[width=1\linewidth]{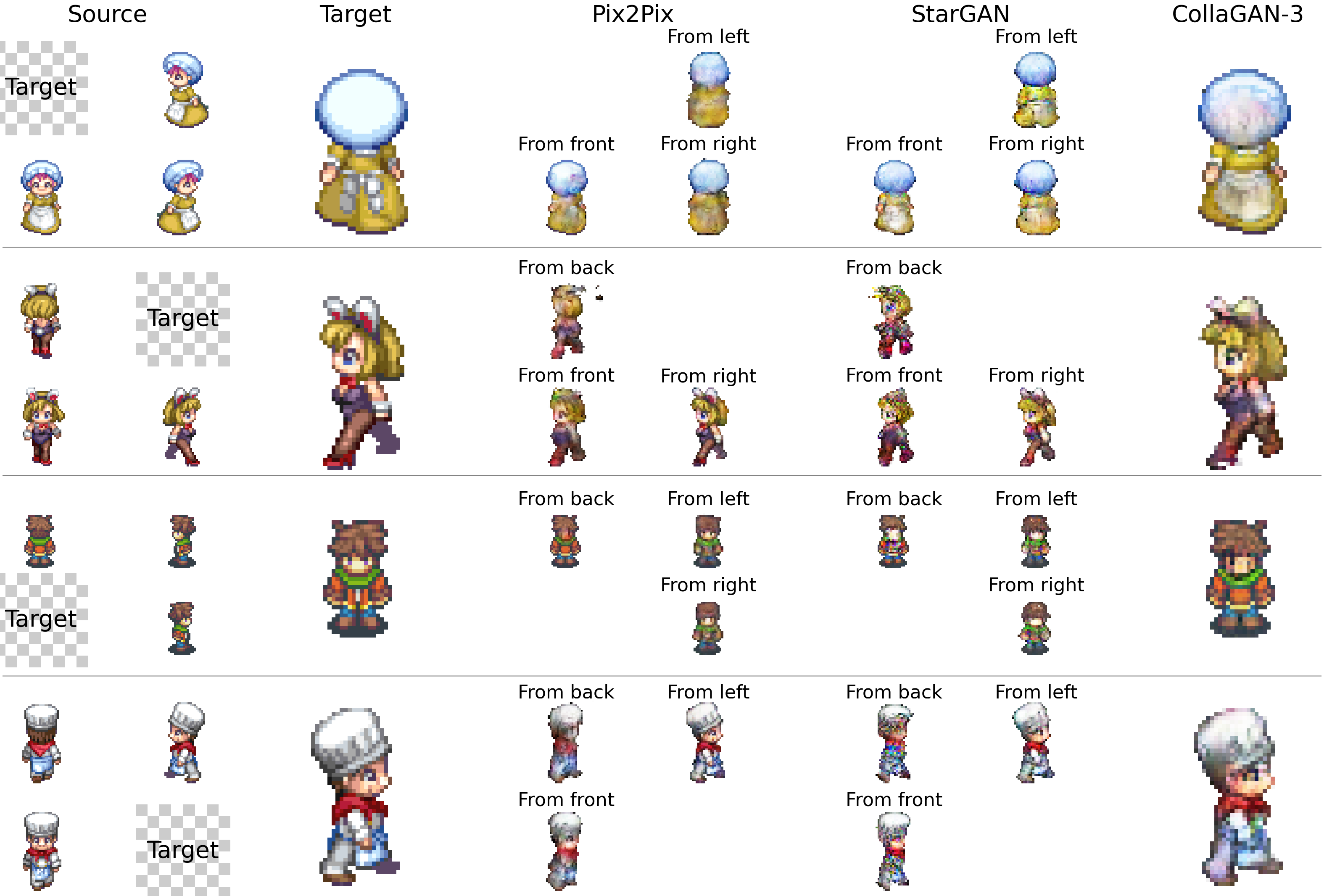}
    \caption{Example images generated in different target domains. The columns show the source images, the target, the generation with the baselines using different source domains, and the generation using all sources with CollaGAN.}
    \Description[Four example characters generated using the baselines Pix2Pix, StarGAN and CollaGAN]{The figure shows four rows, each depicting an example character and a target direction in which we want to generate it. The first column shows the three images of the character in the source poses. The second column shows the target image. The third column shows three images generated by Pix2Pix coming from each specific source pose. The fourth column is similar to the previous one, but StarGAN generated the images. The fifth column shows the single image generated by CollaGAN using all the available sources.}
    \label{fig:collagan-3-vs-baseline}
\end{figure*}

Regarding FID, CollaGAN-3 had the lowest (best) values of the evaluated models in all target poses and, hence, on average too: 1.508 (CollaGAN-3) versus 2.288 (StarGAN) and 4.091 (Pix2Pix). Also, the $L_1$ distance for CollaGAN was the lowest (best), with the averages: 0.04078 (CollaGAN-3), 0.05273 (Pix2Pix), and 0.06577 (StarGAN). Next, we visually analyze the generated images.

Figure~\ref{fig:collagan-3-vs-baseline} shows examples of generated images using the different models, with each row having a different domain as the target. The columns for Pix2Pix and StarGAN show three images per row. As they are models that take a single image as input, we depict the image generated for the target pose from each of the other domains. In contrast, CollaGAN uses all the other domains as input and, hence, has a single generated image for each row.

The quality of the generated images varies with the model and the target pose. We analyze the results qualitatively according to the use of colors and the generated shape. Regarding the former, all models generate images with colors in meaningful positions but employ many variations of the same tones instead of restricting to a small palette. Such undesired behavior can be attenuated by quantizing the colors to the palette of the input images, such as done in~\citep{Coutinho2024PixelGANs} in a post-processing step. 

Regarding the shape, the poses imputed by CollaGAN are very close to the intended, and so are the ones generated by the baseline models that translate images from left to right and vice-versa. Generating images in that scenario usually consists of learning a horizontal flip transformation, which is an easier task endorsed by the lower FID and $L_1$ values when the target is left or right. When the target is back, a noticeable artifact is the faint presence of details from the character's face, especially prominent in the images generated with CollaGAN.



Visually inspecting the results shows that the quality of the images generated by our model is either on par or better than the baselines. We highlight that the CollaGAN-based architecture contains 104,887,616 trainable parameters, which is 22\% smaller than StarGAN and 70\% than the collective Pix2Pix. Next, we assess how the model performs when less images are available.

\subsection{Generating from Fewer Domains}
Even though we propose a model to impute a single missing domain, we also evaluate it in scenarios where it receives two (CollaGAN-2) or only one image (CollaGAN-1). The metrics' values are averaged among all targets and all available sources for each model and scenario (i.e., CollaGAN-3, 2, and 1). 

Table~\ref{tab:fewer-domains-generation} compares the proposed model in those situations. We can observe that both FID and $L_1$ metrics progressively improve as the number of available domains increases, with CollaGAN-2 still having better $L_1$ than Pix2Pix and StarGAN.

\begin{table}[t]
\centering
\caption{FID and $L_1$ metrics in the scenarios of receiving three, two, and one images and the baselines}
\label{tab:fewer-domains-generation}
\begin{tabular}{@{}lrr@{}}
\toprule
Model/Sources & Average FID & Average $L_1$ \\ \midrule
Pix2Pix       & 4.091       & 0.05273       \\
StarGAN       & 2.288       & 0.06577       \\
CollaGAN-1    & 8.393       & 0.06449       \\ 
CollaGAN-2    & 4.277       & 0.05035       \\
CollaGAN-3    & 1.508       & 0.04078       \\ \bottomrule
\end{tabular}
\end{table}

\subsection{Input Dropout}
We evaluated the impact of different batch selection strategies on presenting examples to the proposed model: Should it always see the three available domains, or should they sometimes be omitted?

We investigated always showing all available domains (none), the original input dropout strategy proposed in~\citep{Lee2019CollaGAN:Imputation}, a curriculum learning approach suggested by~\citep{Sharma2019MissingNetwork}, and our proposed conservative tactic. The original approach has an equal chance of presenting three, two, or a single image in a training step. The curriculum learning approach starts training with easier tasks (using three images) and progressively makes it harder (using a single input) until half of the training, then it randomly chooses between the number of domains to drop out for the second part. Lastly, the conservative approach randomly selects the number of images to drop, but with higher probabilities to keep more images: 60\% with 3 images, 30\% with 2, and 10\% with a single image.


\begin{table}[t]
\centering
\caption{FID and $L_1$ of different input dropout strategies when the model receives 3, 2 or 1 images as input}
\label{tab:input-dropout-results}
\begin{tabular}{@{}llrrrr@{}}
\toprule
                            & Sources    & None     & Original          & Curric.   & Conserv.      \\ \midrule
\multirow{3}{*}{FID}        & CollaGAN-3 &  4.816   &  1.911            &  2.160    & \tb{1.508}    \\
                            & CollaGAN-2 & 19.050   &  6.835            &  9.233    & \tb{4.277}    \\
                            & CollaGAN-1 & 32.676   & 11.162            & 20.303    & \tb{8.393}    \\ \midrule
                            & Average    & 18.847   &  6.636            & 10.566    & \tb{4.726}    \\ \midrule
\multirow{3}{*}{$L_1$}      & CollaGAN-3 & 0.04523  & 0.04277           & 0.04222   & \tb{0.04078}  \\
                            & CollaGAN-2 & 0.08003  & 0.05053           & 0.07389   & \tb{0.05035}  \\
                            & CollaGAN-1 & 0.12820  & \tb{0.06243}      & 0.12232   & 0.06449       \\ \midrule
                            & Average    & 0.08449  & 0.05191           & 0.07948   & \tb{0.05187}  \\ \bottomrule
\end{tabular}
\end{table}

Table~\ref{tab:input-dropout-results} presents the results from the models trained with the different input dropout strategies (columns) in the scenarios of having three, two, or one available image as input (rows). We can observe that using any input dropout yields better results than always showing all domains (none). Compared to the original and curriculum learning strategies, our proposed conservative tactic has better FID and $L_1$ metrics on the average of the three scenarios. In particular, regarding FID, the model trained with the conservative input dropout worsens its performance less drastically with the decrease of input domains. Regarding $L_1$, its metrics are better than the other models when two and three images are available.

\subsection{Ablation Study}
To understand the impact of our changes to the original CollaGAN architecture, we trained and evaluated models that progressively added each modification. Table~\ref{tab:ablation} shows the FID and $L_1$ values of the generated images averaged over all domains and among the scenarios of the model receiving three, two, and one input domains. The rows show the results of each modification cumulatively: the first one is the original CollaGAN model without any of our proposed changes, the second introduces the first modification, the third uses two changes, and the last includes all three (our final model).

\begin{table}[t]
\caption{Performance of the modifications made to the original CollaGAN architecture}
\label{tab:ablation}
\begin{tabular}{@{}lrrrr@{}}
\toprule
\multirow{2}{*}{\makecell{Modification\\ (cumulative)}} & \multicolumn{2}{c}{Average FID} & \multicolumn{2}{c}{Average $L_1$} \\ \cmidrule(lr){2-3} \cmidrule(l){4-5} 
                            & Value     & Improv.   & Value     & Improv.   \\ \midrule
Original                    &  8.866    & ---       & 0.06069   & ---       \\
+ Increased capacity        & 11.078    &-24.95\%   & 0.05666   &  6.64\%    \\
+ Forward Replacer          &  6.636    & 25.15\%   & 0.05191   & 14.47\%   \\
+ Conservative Inp. Drop.   &  4.726    & 46.70\%   & 0.05187   & 14.53\%    \\ \bottomrule
\end{tabular}
\end{table}

The original model had 6,565,712 trainable variables, but with the increased capacity, there are 104,887,616 parameters. That change alone improved $L_1$ but worsened FID. The replacement strategy of substituting only the original target with the image generated in the forward step improves both metrics' results. Lastly, training with the proposed conservative input dropout further enhances the results, with FID and $L_1$ values that are 46.7\% and 14.53\% better than the original architecture.

\section{Final Remarks}
We posed the task of generating pixel art characters as a missing data imputation problem and approached it using a deep generative model. It is based on the CollaGAN architecture, from which we proposed changes involving a capacity increase, a conservative input dropout strategy, and a different replacement tactic during the backward step of the training procedure. The experiments showed that all of the changes contributed to achieving better results.

Compared to the baseline models, our approach produces images with similar or better quality when using three domains as input. The model can still produce feasible images in scenarios with fewer available images but with increasingly lower quality. 

In future work, we propose the study of other missing image imputation architectures to the same task tackled here, such as ReMIC~\citep{Shen2021Multi-DomainData} and MM-GAN~\citep{Sharma2019MissingNetwork}. Differently from CollaGAN, both methods can receive and generate images in any number of domains. Another line of investigation is to approach the task with architectures that disentangle the source images into content and style codes~\citep{Huang2018MultimodalTranslation} and also latent diffusion models~\citep{Rombach2021High-ResolutionModels}. An interesting outcome of such architectures is their multi-modality nature, in that they can generate different suggestions for the same input.

\section{Acknowledgments}
\hladd{This work was partially supported by CAPES, CNPq and Fapemig.}

\bibliographystyle{ACM-Reference-Format}
\bibliography{references}

\end{document}